\documentclass{elsarticle}
\makeatletter
\def\ps@pprintTitle{%
  \let\@oddhead\@empty
  \let\@evenhead\@empty
  \def\@oddfoot{}%
  \let\@evenfoot\@oddfoot}
\makeatother

\usepackage{amssymb}
\usepackage{amsmath}
\usepackage{url}
\usepackage{makecell}
\usepackage{multirow}
\usepackage{tabularx} 
\usepackage{ragged2e} 
\usepackage{booktabs} 
\usepackage{cleveref}
\usepackage{listings}
\def\figurecrefname{Fig.}
\crefformat{figure}{#2\figurecrefname~#1#3}

\def\tablecrefname{Tab.}
\crefformat{table}{#2\tablecrefname~#1#3}

\def\eauationcrefname{Eq.}
\crefformat{equation}{#2\eauationcrefname~(#1)#3}

\usepackage[caption=false,font=footnotesize,labelfont=rm,textfont=rm,subrefformat=parens]{subfig}

\usepackage{caption}
\begin{document}

\begin{frontmatter}



\title{KAConvNet: Kolmogorov–Arnold Convolutional Networks for Vision Recognition}

\cortext[cor1]{Corresponding author}

\author[1,2]{Zhaoxiang Liu} 
\author[1,2]{Zhicheng Ma\corref{cor1}}
\ead{mazhicheng99@163.com}
\author[1,2]{Kaikai Zhao}
\author[1,2]{Kai Wang}
\author[1,2]{Shiguo Lian\corref{cor1}}
\ead{liansg@chinaunicom.cn}

\affiliation[1]{organization={Data Science \& Artificial Intelligence Research Institute},
            addressline={China Unicom}, 
            city={Beijing},
            country={China}}
            
\affiliation[2]{organization={Unicom Data Intelligence},
            addressline={China Unicom}, 
            city={Beijing},
            country={China}}


\begin{abstract}
The Convolutional Neural Networks (CNNs) have been the dominant and effective approach for general computer vision tasks. Recently, Kolmogorov–Arnold neural networks (KANs), based on the Kolmogorov-Arnold representation theorem, have shown potential to replace Multi-Layer Perceptrons (MLPs) in deep learning. KANs, which use learnable nonlinear activations on edges and simple summation on nodes, offer fewer parameters and greater explainability compared to MLPs. However, there has been limited exploration of integrating the Kolmogorov-Arnold representation theorem with convolutional methods for computer vision tasks.
Existing attempts have merely replaced learnable activation functions with weights, undermining KANs' theoretical foundation and limiting their potential effectiveness. Additionally, the B-spline curves used in KANs suffer from computational inefficiency and a tendency to overfit.
In this paper, we propose a novel Kolmogorov–Arnold Convolutional Layer that deeply integrates the Kolmogorov-Arnold representation theorem with convolution. This layer provides stronger method interpretability because it is based on established mathematical theorems and its design has theoretical alignment. Building on the Kolmogorov–Arnold Convolutional Layer, we design an efficient network architecture called KAConvNet, which outperforms existing methods combining KAN and convolution, and achieves competitive performance compared to mainstream ViTs and CNNs. We believe that our work offers valuable insight into the field of artificial intelligence and will inspire the development of more innovative CNNs in the 2020s. The code is publicly available at https://github.com/UnicomAI/KAConvNet.
\end{abstract}




\end{frontmatter}



\section{Introduction}
The advancement of deep learning architectures has accelerated progress in computer vision. Since LeNet \cite{lecun1998gradient} introduced convolution neural networks (CNNs) into image recognition tasks, CNNs have rapidly evolved and now dominate various vision tasks, including classification, detection, and segmentation. 
Following this, ViT \cite{dosovitskiy2020image} expanded Transformer \cite{vaswani2017attention} from natural language processing to computer vision for the first time, and researchers have continually improved upon it.

Kolmogorov–Arnold Neural Networks (KANs) \cite{liu2024kan} have recently emerged as a promising alternative to Multi-Layer Perceptrons (MLPs) in deep learning. Unlike MLPs, KANs employ learnable nonlinear activations on edges and simple summation at nodes, resulting in fewer parameters and enhanced fitting capabilities. Despite their potential, limited research has explored the integration of KANs with convolutional neural networks (KACNNs) for computer vision tasks.
\begin{figure}[!ht]
    \centering
    \includegraphics[width=0.8\textwidth]{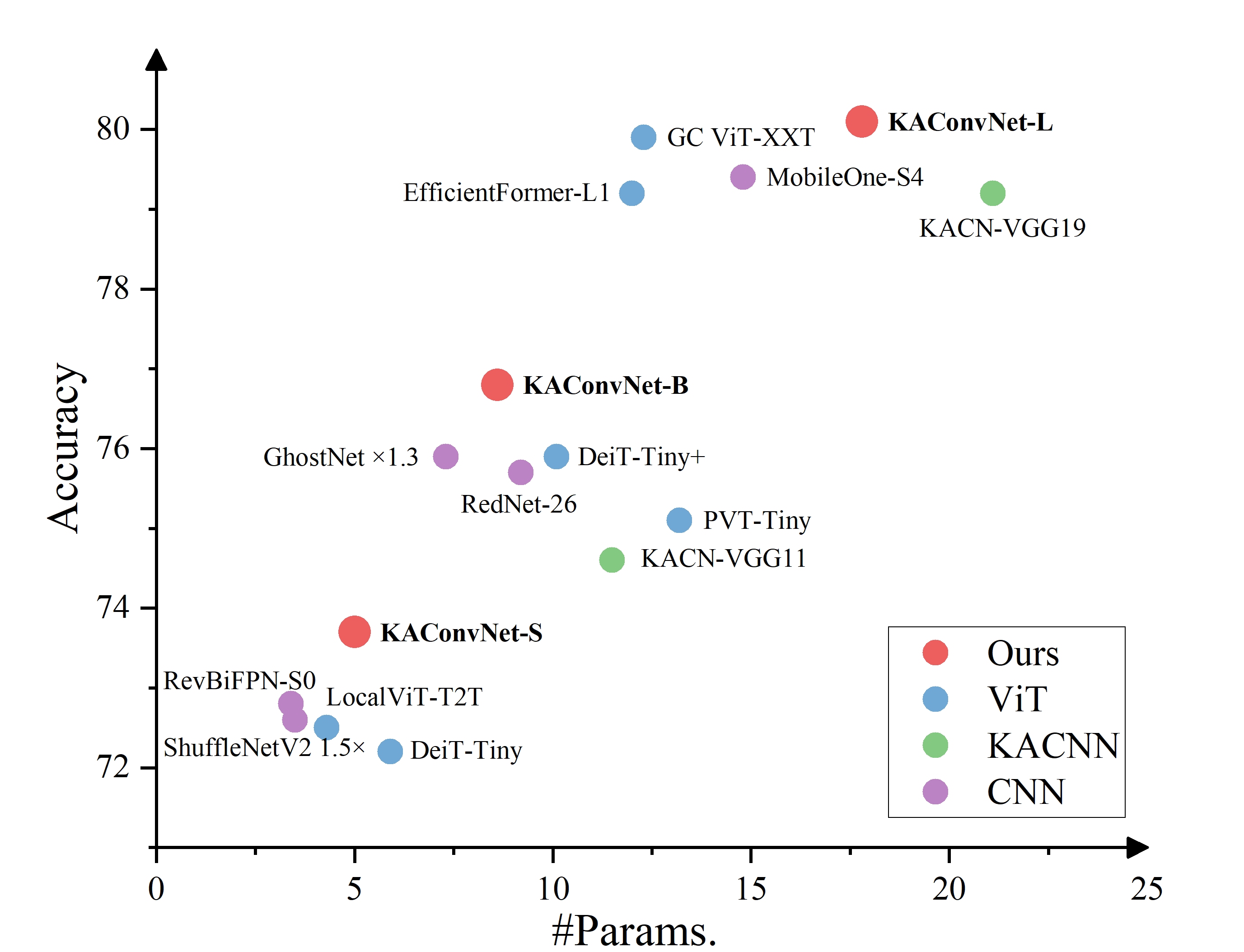}
    \caption{Comparison of some competitive methods on ImageNet-1K. Our KAConvNets achieve higher accuracy with fewer parameters.}
    \label{fig:firsfig}
    \vspace{-0.1cm}
\end{figure}

One notable attempt is CKAN \cite{bodner2024convolutional}, which incorporates KANs into convolution layers. However, it demonstrates subpar performance on the MNIST dataset. And CKAN employs a shared learnable activation function across all channels, restricting the ability of individual channels to express unique features. Another study, KACN \cite{drokin2024kolmogorov}, applies the learnable activation function across all channels, resulting in a significant increase in the number of parameters.
Both CKAN and KACN deviate from the original Kolmogorov–Arnold representation theorem by merely placing the learnable activation function on the weights. This simplified approach compromises the theoretical foundation and potential effectiveness of KANs in CNN-based architectures.

\begin{figure}[t]
\centering
\subfloat[Training Loss]{\includegraphics[width = 0.45\textwidth]{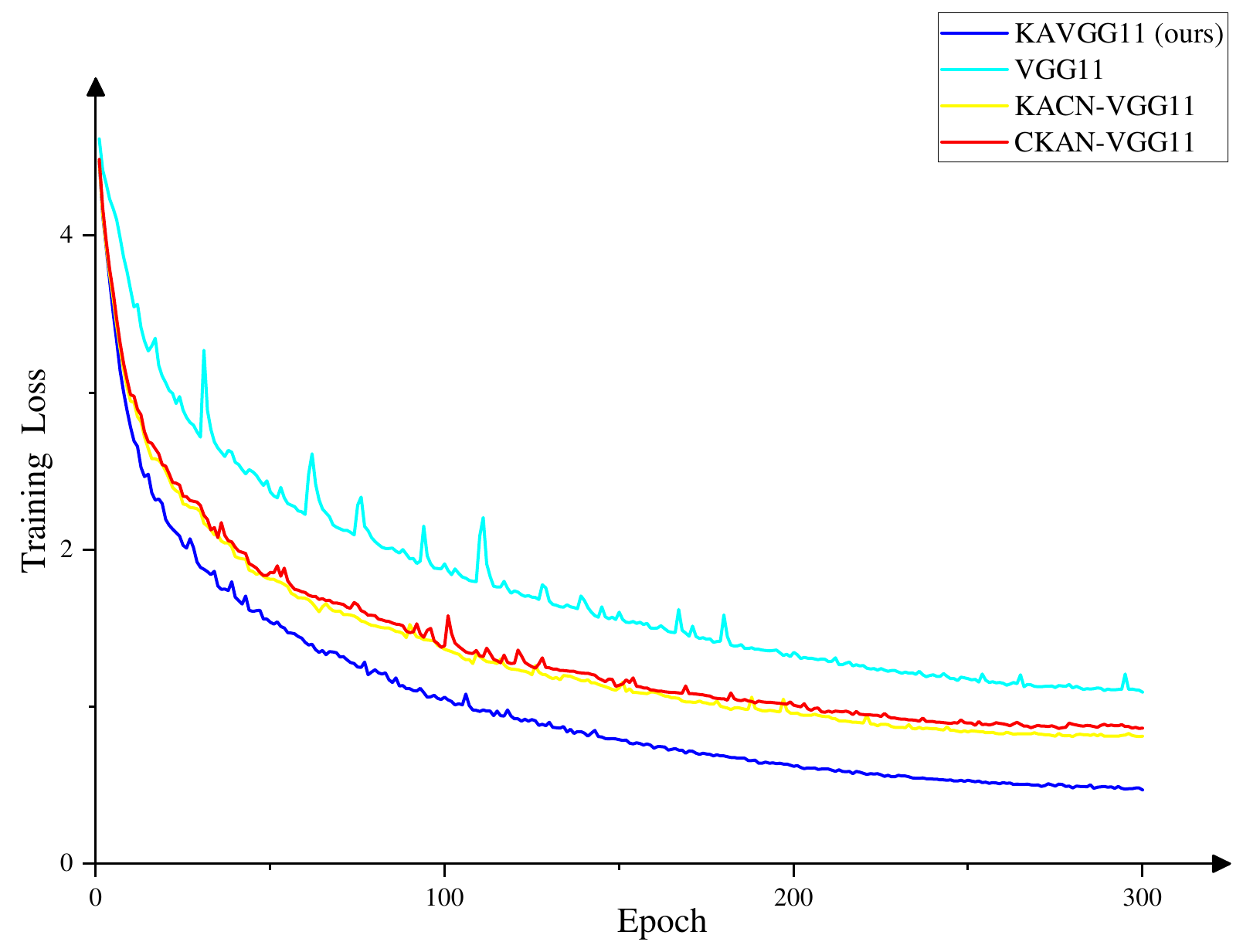}\label{subfig:trainloss}}
\vspace{-2.5mm}
\subfloat[Test Accuracy]{\includegraphics[width = 0.45\textwidth]{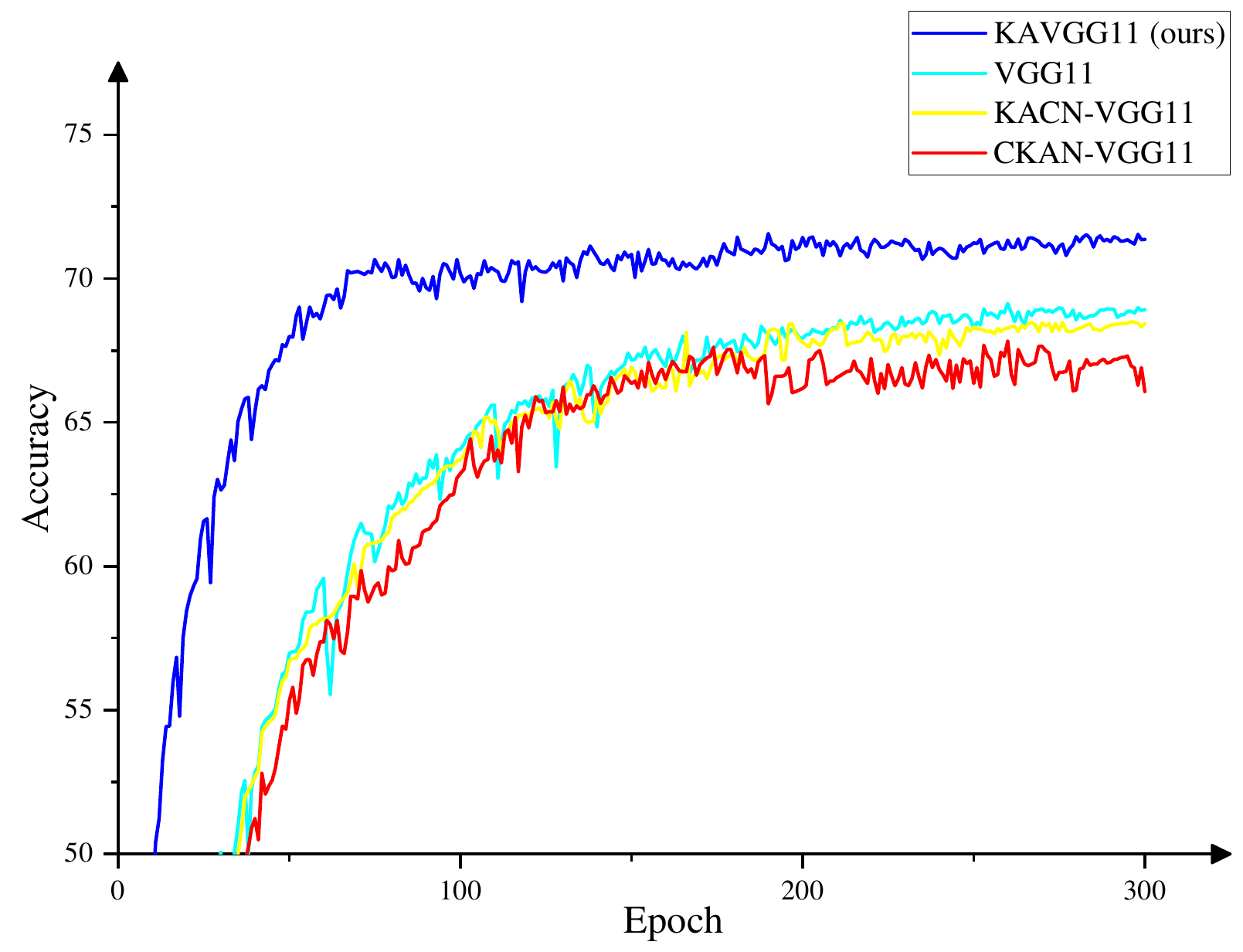}\label{subfig:acc}}
\caption{The (a) training loss and (b) test accuracy after 300 epoch on the CIFAR-100. VGG-like CKAN and VGG-like KACN which adopt B-spline overfit compared to convolution and our method.}
\label{fig:overfit}
\vspace{-2.5mm}
\end{figure}

In addition, the B-spline function used in CKAN and KACN, while effective for fitting certain functions, introduces two key challenges. First, parameter updates for B-splines slow down training, especially with finer interval divisions and higher-order splines. Second, B-splines are prone to overfitting in computer vision tasks, leading to lower test accuracy despite reduced training loss, as shown in \cref{fig:overfit}.

In this paper, we propose a novel and efficient Kolmogorov–Arnold Convolution Layer (KAConvLayer) by deeply integrating KANs with convolution layers. Our approach applies the inner function of the Kolmogorov–Arnold representation theorem to each channel for information aggregation within each channel, and the outer function across all channels for effective information exchange. 
This design allows each channel to maintain distinct functional mappings while reducing the total number of input elements and learnable functions. Consequently, our method achieves faster computation and convergence, while enhancing representational flexibility compared to existing approaches.
In view of the fact that B-spline curves are prone to overfitting, we use a piecewise linear learnable activation function on the convolution layer nodes to prevent overfitting.
Based on this layer, we design a new network architecture called KAConvNet, which demonstrates outstanding performance in computer vision tasks such as classification, detection, and segmentation, as illustrated in \cref{fig:firsfig}.
Finally, we explore the advantages and limitations of KAConvLayers, providing valuable insights for their application in computer vision and laying a solid foundation for future research on KAN-CNN integration. The core contribution of our work is to propose an advanced KAN-CNN hybrid paradigm that delivers significant and consistent improvements in both accuracy and inference speed over existing KAN-based methods. 

\section{Related Work}
In this section, we summarize the latest applications of Kolmogorov–Arnold representation theorem in deep learning and briefly review the development of convolutional neural networks.

\subsection{Kolmogorov–Arnold Neural Networks}
Vladimir Arnold and Andrey Kolmogorov established that if $f$ is a multivariate continuous function on a bounded domain, and then the $f$ can be expressed as a finite composition of continuous functions of one-dimensional variable and the binary operation of addition \cite{kolmogorov1961representation,braun2009constructive}.
That is to say that a smooth $f(\cdot)$ can be written as
\vspace{-0.1cm}
\begin{equation}
f(x_{1},\ldots,x_{n})=\sum_{q=0}^{2n} \Phi_{q}\Biggl(\sum_{p=1}^{n}\psi_{q,p}(x_{p})\Biggr) \\
\vspace{-0.2cm}
\end{equation}

However, it is unclear how to choose the outer and inner functions $\Phi_{q}$ and $\psi_{q,p}$. 
KAN \cite{liu2024kan} creatively uses neural networks to approximate these functions and places learnable activation functions on weights of MLP.
Extensive numerical experiments show that the KAN is a powerful new neural network architecture known for its improved performance and interpretability.

Inspired by KAN, TKANs \cite {genet2024tkan} combines the strength of both KAN and LSTM, and is composed of Recurring Kolmogorov-Arnold Networks (RKANs) Layers embedding memory management to perform multi-step time series forecasting with enhanced accuracy and efficiency.
PointNet-KAN \cite{kashefi2024pointnet} retains the core principle of PointNet by using shared KAN layers and applying symmetric functions for global feature extraction, ensuring permutation invariance with respect to the input features. 
KAN-CUT \cite{Mahara2024KAN-CUT} replaces the two-layer MLP with a two-layer KAN in the existing model, which substitution favors the generation of more informative features in low-dimensional vector representations and can utilize more effectively to produce high-quality images in the target domain. 
U-KAN \cite{li2025ukan} integrates the dedicated KAN layers on the tokenized intermediate representation based on UNet.

Because Fourier coefficients are more dense than B-splines and are easier to optimize, FourierKAN \cite{fourierkan2024} replaces the activation function of B-spline with Fourier coefficients.
Building on this, Wav-KAN \cite{bozorgasl2024wav} replaces the Fourier coefficients with wavelet coefficients. This approach efficiently handles both high-frequency and low-frequency components of the input data.
GKAN \cite{kiamari2024gkan} adapts the KAN principles to graph-structured data. It introduces learnable spline-based functions between layers, revolutionizing information processing within graph structures. 
KAL-Net \cite{kalnet2024} presents a novel architecture that utilizes Legendre polynomials. Specifically, KAL-Net employs Legendre polynomials up to a certain order for input normalization, which allows for more efficient capture of nonlinear relationships. 
JacobiKAN \cite{JacobiKAN2024} replaces B-splines with Jacobi polynomials.
Fast-KAN \cite{FastKAN2024} substitutes the 3rd-order B-spline basis in original KANs with Radial Basis Functions (RBFs). 
CKAN \cite{bodner2024convolutional} employs a shared learnable activation function across all channels, restricting the ability of individual channels to express unique features and leading to subpar performance on the MNIST dataset. KACN \cite{drokin2024kolmogorov} applies the learnable activation function across all channels but makes the the number of parameters increase.

Despite these advancements, networks based on KAN often suffer from slow speeds and long training times.
Efficient-KAN \cite{efficientkan2024} reformulates the computation by activating the input with different basis functions and then combining them linearly to reduce the memory cost and make the computation a straightforward matrix multiplication.

In constrast, we propose a novel and efficient Kolmogorov–Arnold Convolution Layer (KAConvLayer) by deeply integrating KANs with convolution layers. Our approach applies the inner function of the Kolmogorov–Arnold representation theorem to each channel for information aggregation within each channel, and the outer function across all channels for effective information exchange. This design allows each channel to maintain distinct functional mappings while reducing the total number of input elements and learnable functions. Consequently, our method achieves faster computation and convergence, while enhancing representational flexibility compared to existing approaches.

\subsection{Convolution Neural Networks}

LeNet \cite{lecun1998gradient} is the first CNN designed for image recognition tasks, providing crucial insights for the development of subsequent CNNs. AlexNet \cite{krizhevsky2012imagenet}, consisting of five convolution layers and three fully connected layers, achieves top results in the ILSVRC-2012 competition. VGG \cite{simonyan2014very} reduces the size of the convolution kernels while increasing the number of network layers, establishing the practice of using stacks of small convolution kernels in CNNs.

GoogLeNet introduces inception modules, enabling multi-scale processing within the network and improving both depth and width without increasing computational costs. ResNet \cite{he2016deep} addresses the vanishing gradient problem by adding skip connections to learn residual functions relative to the layer inputs, facilitating the optimization of very deep networks.

Several works focus on proposing new convolution layers. Dilation convolutions \cite{yu2015multi} systematically aggregate multi-scale contextual information without losing resolution. MobileNet \cite{howard2017mobilenets} and Xception \cite{chollet2017xception} use depthwise separable convolutions, performing depthwise convolution followed by pointwise convolution, which reduces theoretical computation. Deformable convolution \cite{dai2017deformable} introduces 2D offsets to the regular grid sampling locations in standard convolution, allowing free-form deformation of the sampling grid.

In the 2020s, the Vision Transformer demonstrates exceptional performance in computer vision, prompting researchers to explore modern CNN architectures. ConvNeXt \cite{liu2022convnet} is the first to design a modern architecture inspired by the Vision Transformer. 
RepVGG \cite{ding2021repvgg} introduces structural re-parameterization to decouple a training-time multi-branch topology from an inference-time plain architecture, highlighting the potential of multi-branch re-parameterized CNNs and inspiring subsequent large-kernel works. RepLKNet \cite{ding2022scaling} scales up the kernel size to $31 \times 31$ by re-parameterizing a small kernel (e.g., $5\times5$) parallel to it, showcasing the versatility of re-parameterized CNNs in achieving enhanced performance. SLaK \cite{liu2022more} combines decomposed convolution with dynamic sparsity to scale up kernels to $51 \times 51$ in the form of two rectangular convolutions. UniRepLKNet \cite{ding2024unireplknet} employs multi-branch dilated convolutions to enhance the capability of large kernels in capturing sparse patterns and introduces the re-parameterization method of dilated convolution.

\begin{figure}[!t]
\centering
\subfloat[An illustration of KAConvLayer. The red boxes and lines denote the internal functions $\phi$ of Kolmogorov-Arnold representation theorem, and the green lines denote the outer functions of Kolmogorov-Arnold Representation theorem. The blue lines denote the fully connected layer, which are used to adjust the number of channels.]{\includegraphics[width = 0.909\textwidth]{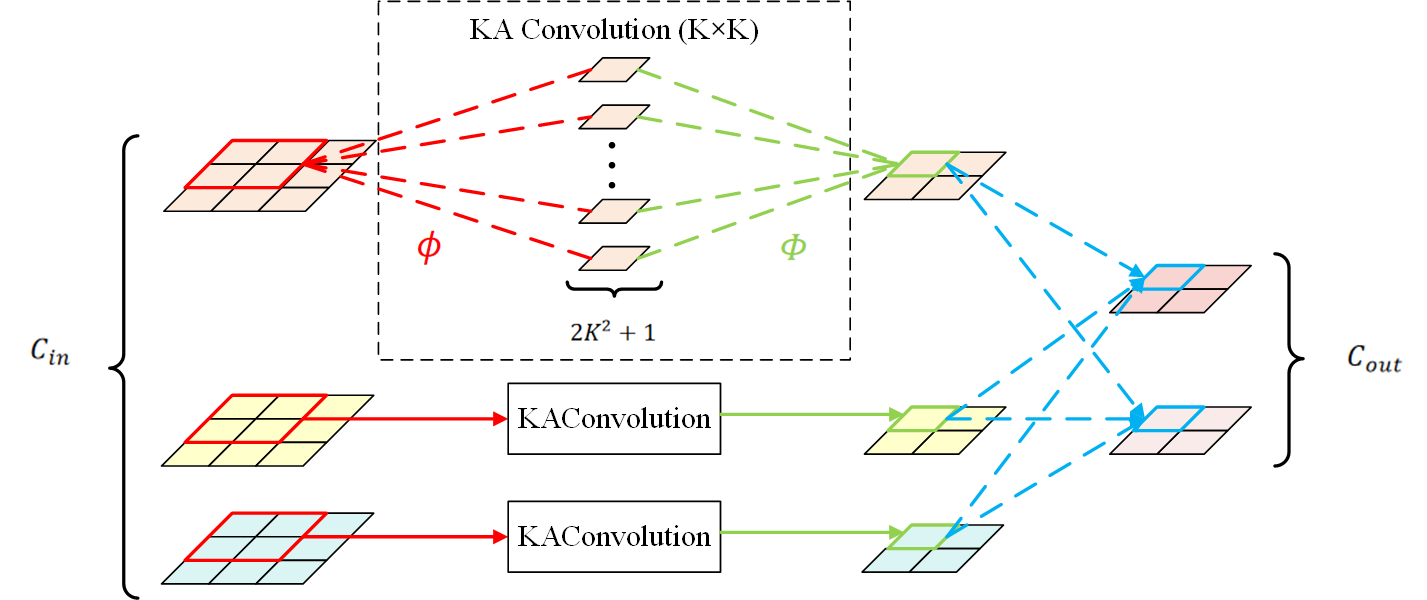}\label{kaconv}}
\\
\subfloat[Activation functions in fully connected layers and convolutions at different positions.]{\includegraphics[width = 0.9\textwidth]{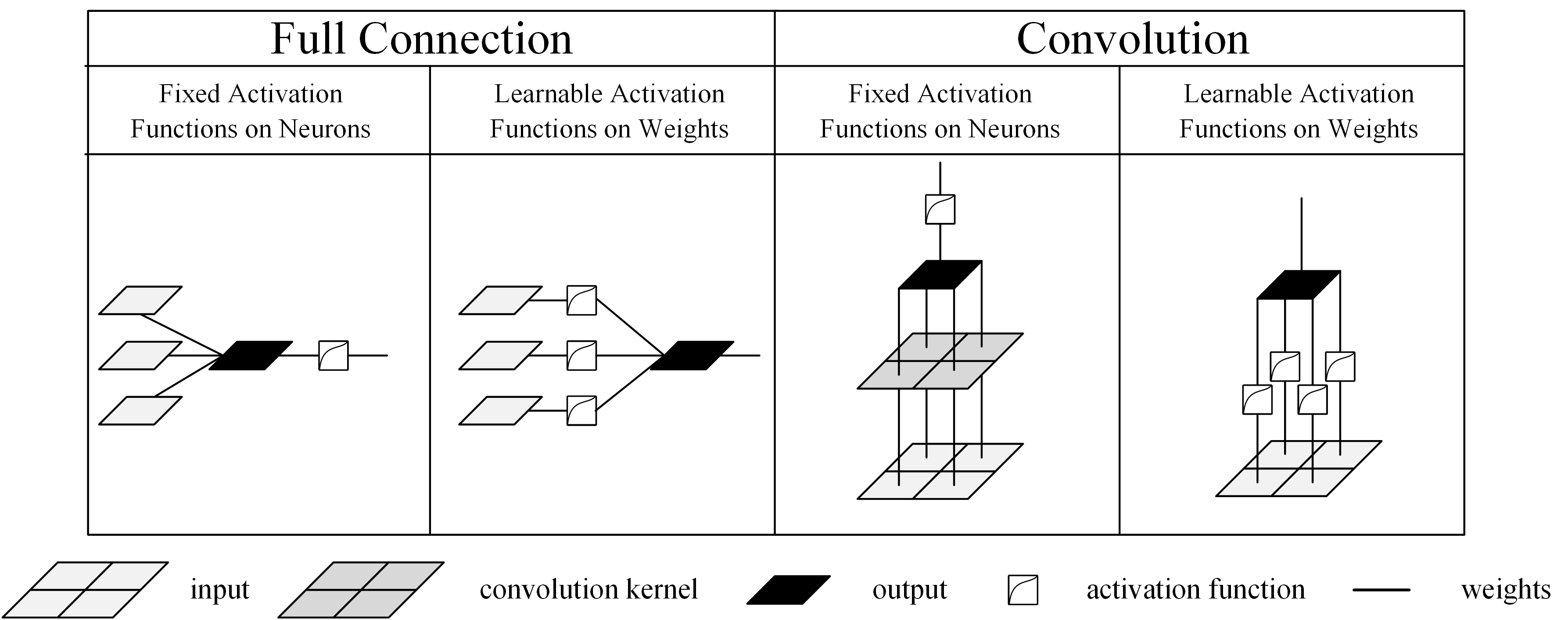}\label{actondifferent}}
\caption{Illustrations of (a) KAConvLayer structure and (b) sketch of activation on different positions.}
\label{actandkaconv}
\end{figure}

\section{Kolmogorov–Arnold Convolution Neural Network}
In this section, we first revisit the Kolmogorov–Arnold representation theorem and propose a new Kolmogorov–Arnold Convolution Layer (KAConvLayer) by applying this theorem to convolution. Then we propose KAConvNet, a pure CNN architecture with KAConvLayer design. 
\subsection{Kolmogorov–Arnold Convolution Layer}
Given an image patch $I_{K\times K}$ with a size of $K\times K$ and an unknown multivariate continuous function $f(\cdot)$, by Kolmogorov-Arnold representation theorem representation theorem we can get
\vspace{-0.1cm}
\begin{equation}
\begin{split}
f(I_{k\times k})=&f(\begin{bmatrix}
  I_{1,1}& I_{1,2} &\cdots  & I_{1,K}\\
  I_{2,1}& I_{2,2} & \cdots & I_{1,K}\\
  \vdots& \vdots &\ddots  & \vdots\\
  I_{K,1}& I_{K,2} & \cdots &I_{K,K}
\end{bmatrix}) 
\\ 
=& \sum_{q=1}^{2K^2+1} \Phi_q(\sum_{i=1}^{K}\sum_{j=1}^{K} \phi _{i,j}^{q}(I_{i,j}))
\end{split}
\label{2DkAequation}
\vspace{-0.2cm}
\end{equation}
where $i$ and $j$ denote pixel coordinates of image patch, and the $\Phi_q$ and $\phi _{i,j}^{q}$ represent the outer functions and internal functions respectively. Specifically, the single internal function extracts simple local features within a pixel receptive field, while the outer function aggregates multiple such local features to form complex hierarchical feature patterns. 

\cref{2DkAequation} implies that function $f(\cdot)$ can be obtained by identifying suitable individual functions $\Phi_q$ and $\phi _{i,j}^{q}$. 
This insight motivates the development of a neural network that explicitly incorporates the parameterization defined in \cref{2DkAequation}. Since all functions to be learned are univariate functions, we can parametrize each 1D function as rectified learnable activation functions.
The rectified learnable function $\phi(\cdot)$ is the product of a basis function $b(\cdot)$ and a learnable function $\tilde{\phi}(\cdot)$, so the $\phi(\cdot)$  be described as follows
\begin{equation}
    \phi(x) = w_{1} b(x) \times w_{2} \tilde{\phi}(x)
\label{wx1wx2}
\end{equation}
where $w_{1}$ and $w_{2}$ denote weights. We can multiply the two different activation functions to achieve \cref{wx1wx2}.
And we select
\vspace{0.1cm}
\begin{equation}
    b(x) = SiLU(x)=\frac{x}{1+e^{-x}} 
\end{equation}
in most cases. 
B-spline activations, as higher-order polynomial functions, possess strong representational power but can also exhibit excessive flexibility, potentially leading to the modeling of high-frequency noise and an increased risk of overfitting due to their higher model capacity.
So we design a new learnable function named Grid Linear (GLinear) function as $\tilde{\phi}(x)$, the function can be formulated as
\begin{equation}
\begin{split}
&GLinear(x) 
=\begin{cases}
& \beta + \alpha_{1}x, \qquad \qquad \qquad \qquad \qquad  \qquad x\leq g_{1} \\
& GLinear(g_{i-1}) + \alpha_{i}(x-g_{i - 1}),  \quad g_{i-1}<x\leq g_{i} \quad(i \geq 2)\\
& GLinear(g_{n}) + \alpha_{n+1}(x-g_{n}), \qquad \quad g_{n}\leq x \\
\end{cases}
\end{split}
\end{equation}
where $\{\alpha_i\} (i=1,\ldots,n,n+1)$ and $\beta$ are learnable parameters, the $G=\{g_1, \ldots, g_{n-1}, g_{n}\}$ ($g_1<\ldots<g_{n-1}<g_{n}$) denotes the interval division.
The piecewise linear GLinear function has lower functional complexity (e.g., lower VC dimension, in an informal sense), which is generally associated with tighter generalization bounds and improved robustness to overfitting.
From a computational perspective, the evaluation of a B-spline of degree k requires $\mathcal{O}(k)$ operations, whereas the GLinear function involves only simple linear interpolation, thus substantially reducing computational cost and improving efficiency. 
The denser the interval division, the better the function fits, but at the same time the larger the number of parameters. We use $n=1, g_1=0$, which is the simplest interval partitioning, as the default value.

\begin{figure}[!t]
    \centering
    \includegraphics[width=0.8\textwidth]{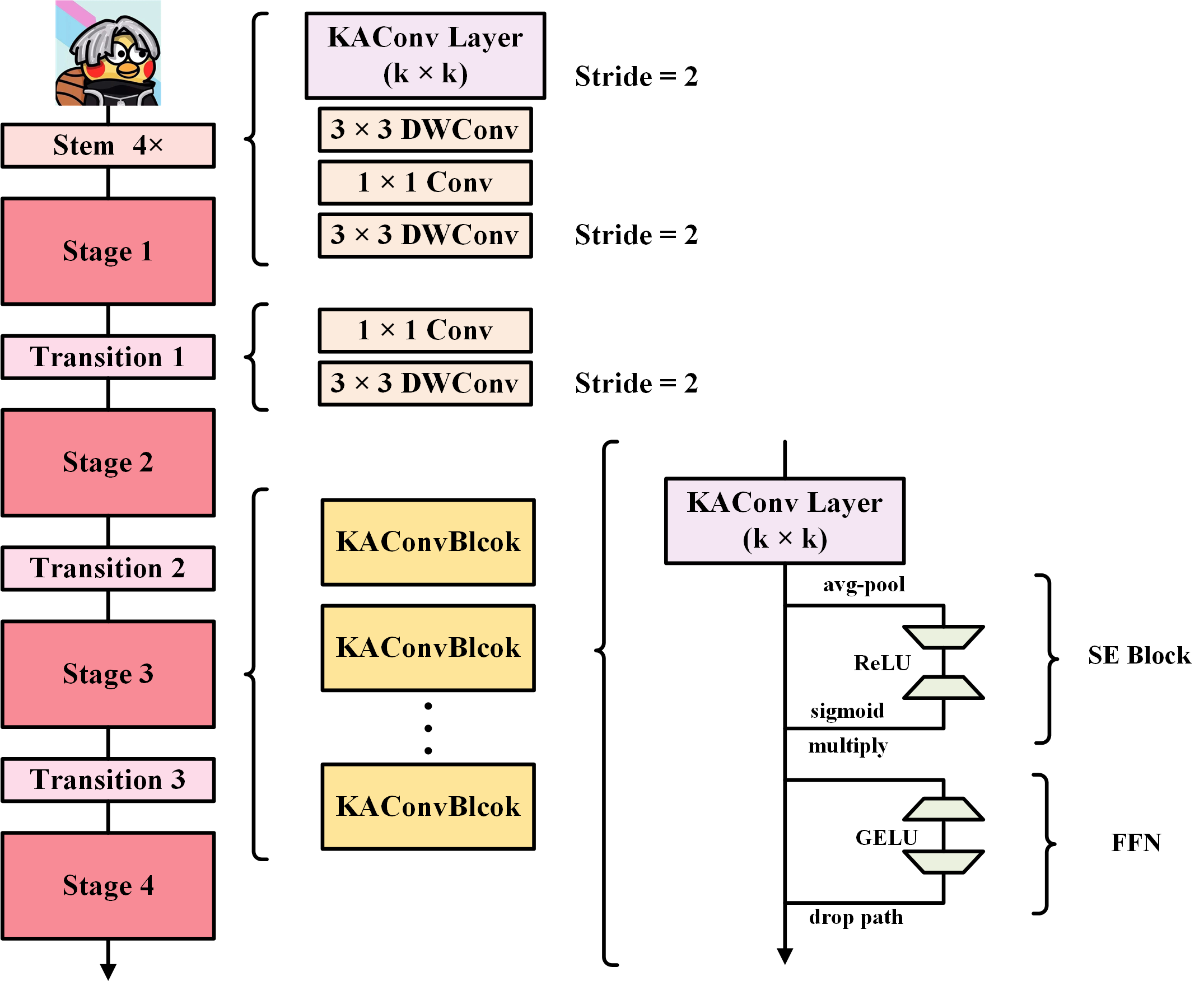}
    \caption{Architectural design of KAConvNet, which comprises Stem, Stages and Transitions as conventional CNNs.}
    \label{fig:kaconvnet}
\vspace{-0.1cm}
\end{figure}

And now, we can use pointwise convolution with learnable activation functions on weights to represent external functions $\Phi_q$ and use convolution with learnable activation functions to represent internal functions $\phi_{i,j}^{q}$ for each input channel of input feature map respectively, as shown in the green and red lines in the dashed box of Figure \subref*{kaconv}. When this calculation process slides over the input data, it is considered convolution. We refer to the entire process as KA convolution, with the patch size $K \times K$ referred to as the kernel size of KA convolution. 
As for the specific algorithm implementation, we use the unfold function and channel-separable rectangular convolution to maximize the acceleration of this process.

In order to generalize the KA convolution to arbitrary channels, we propose the KAConvLayer module as shown in \cref{actandkaconv}. In this module, KA convolutions with KA kernel size $K$ are applied to each input channel in a multi-channel feature map with $C_{in}$ channels. Subsequently, for changing the number of channels to $C_{out}$ and channel mixing, a fully connected layers are followed.
To further explain the concept, Figure \subref*{actondifferent} also highlights the differences between traditional fixed activation functions applied to neurons and the learnable activation functions applied to weights.

\subsection{Network Architecture}

We sketch the architecture of KAConvNet in \cref{fig:kaconvnet}, and adopt a 4-stage framework which is same as the popular CNNs \cite{chen2024pelk,ding2024unireplknet,ding2022scaling}.

\textbf{Stem:} In the initial layers of our model, we focus on achieving high performance in downstream dense-prediction tasks by capturing more details. To accomplish this, we use several convolution layers at the beginning. KAConvNet starts with a $3\times3$ KAConvLayer with $2\times$ downsampling. Following this, it includes a $3\times3$ depthwise convolution layer to capture low-level patterns, a common $1\times1$ convolution layer, and another $3\times3$ depthwise convolution layer for further $2\times$ downsampling. The Stem has a total of $4\times$ downsampling of the input image.

\textbf{Stages:} For stages 1-4, each stage contains several KAConvBlocks, which all use residual structure. The KAConvBlock uses a KAConvLayer and a channel attention named SE Block \cite{hu2018squeeze} to further enhance spatial aggregation transformations and channel mixing. We set the kernel sizes $K$ of KA convolution layers to 3 as default, consistent with kernel sizes used in common convolution layers.

\textbf{Transitions:} Transitions are strategically placed between stages to enhance the model's performance. Initially, we increase the channel dimension using a $1\times1$ convolution layer. This is followed by $2\times$ downsampling achieved through a $3\times3$ depthwise convolution layer.

In summary, our network only includes common convolution, channel attention and KA convolution we proposed.
Each stage has two architectural hyper-parameters: the number of KAConvBlock $B_{i}$ and the channel dimension $C_{i}$.
So a KAConvNet architecture is defined by $[B_1, B_2, B_3, B_4]$ and $[C_1, C_2, C_3, C_4]$.

\section{Experiments}
In this section, we first introduce the implementation details of experimental setup, and the we evaluate our KAConvNet on image classification, object detection and semantic segmentation tasks. Since KAConvLayers use only small-kernel convolutions (i.e., $3 \times 3$), our network contains fewer parameters. Consequently, this paper primarily focuses on comparison with lightweight CNNs and ViTs to demonstrate the superiority of KAConvLayers, rather than optimizing for deployment-oriented trade-offs.
Finally, the proposed network is investigated through several ablation studies.

\subsection{Experimental Setup}

As for datasets, we use ImageNet-1K for image classification, COCO \cite{lin2014microsoft} for object detection, and Cityscapes \cite{cordts2016cityscapes} for semantic segmentation.

By default, we set all KAConvLayers in KAConvNet with $K=3$ as mentioned above. And we name the KAConvNet with $[B_1, B_2, B_3, B_4] $ $= [1, 1, 3, 1]$ and $[C_1, C_2, C_3, C_4]=[32, 64, 128, 256]$ as \textbf{KAConvNet-S}, the one with $[B_1, B_2, B_3, B_4] = [2, 2, 6, 2]$ and $[C_1, C_2, C_3, C_4]=[32, 64, 128, 256]$ as \textbf{KAConvNet-B} and the one with $[B_1, B_2, B_3, B_4] $ $= [2, 2, 6, 2]$ and $[C_1, C_2, C_3,$ $ C_4]=[48, 96, 192, 384]$ as \textbf{KAConvNet-L}.

\begin{table}[!h]
\centering
\footnotesize
\setlength{\tabcolsep}{6mm}
\caption{A comprehensive comparison on the ImageNet-1K classification dataset.}
\vspace{-0.1cm}
\begin{tabular}{lccc}
\hline
Method             & \#Params.& FLOPs   & Acc (\%) \\ \hline
RedNet-101 \cite{li2021involution}         & 25.6M    & 4.1G     & 79.1 \\
MobileOne-S4 \cite{vasu2023mobileone}      & 14.8M    & 2.9G     & 79.4 \\
EfficientFormer-L1 \cite{li2022efficientformer} & 12.3M    & 1.3G     & 79.2   \\
GC ViT-XXT \cite{hatamizadeh2023global}        & 12.0M    & 2.1G     & 79.9  \\
KACN-VGG19 \cite{drokin2024kolmogorov}        & 21.1M    & 5.1G     & 79.2 \\
Fast-KAN-ResNet18 \cite{FastKAN2024}           & 23.8M    & 4.2G     & 79.6   \\
JacobiKAN-ResNet18 \cite{FastKAN2024}           & 22.6M    & 4.2G     & 79.5   \\
VMamba-T\cite{liu2024vmamba}                  & 15.5M    & 2.6G         & 80.0     \\
MambaVision \cite{hatamizadeh2025mambavision}        & 17.8M    & 2.8G          & \textbf{80.1}     \\
\textbf{KAConvNet-L (Ours)} & \textbf{17.5M}    & \textbf{2.9G}     & \textbf{80.1} \\ \hline
GhostNet ×1.3 \cite{han2020ghostnet}     & 7.3M     & 0.2G     & 75.7 \\
RedNet-26 \cite{li2021involution}         & 9.2M     & 1.7G     & 75.9 \\
DeiT-Tiny+ \cite{touvron2021going}        & 10.1M    & 2.1G     & 75.9 \\
KACN-VGG11 \cite{drokin2024kolmogorov}        & 11.5M    & 2.1G     & 74.6 \\
Vim-T \cite{zhu2024vision}                     & 7.0M     & \textbf{---}        & 76.1  \\
\textbf{KAConvNet-B (Ours)} & \textbf{8.6M}     & \textbf{1.4G}     & \textbf{76.8} \\ \hline
ShuffleNetV2 1.5× \cite{ma2018shufflenet} & 3.5M     & 0.3G    & 72.6 \\
RevBiFPN-S0 \cite{chiley2023revbifpn}       & 3.4M     & 0.3G    & 72.8 \\
DeiT-Tiny \cite{touvron2021going}          & 5.9M     & 1.1G    & 72.2 \\
LocalViT-T2T \cite{li2021localvit}       & 4.3M     & 1.2G    & 72.5 \\
\textbf{KAConvNet-S (Ours)} & \textbf{5.0M}     & \textbf{0.7G}    & \textbf{73.7} \\  \hline
\end{tabular}
\label{table:imagenet1k}
\vspace{-1mm}
\end{table}

In the image classification experiments, the initial learning rate is set to 2e-3, with a warm-up strategy and cosine annealing applied to adjust the rate throughout training. Models are trained for 300 epochs using the AdamW optimizer \cite{loshchilov2017decoupledadmw}, with a weight decay of 1e-4 and beta parameters of (0.9, 0.99). The warm-up phase lasts for 5 epochs, and the minimum learning rate under cosine annealing is set to 1e-6. For data augmentation, we employ RandAugment, Mixup, and CutMix, with both Mixup and CutMix using an alpha value of 0.5.

For object detection, the initial learning rate is set to 2.5e-3, and the same warm-up and cosine annealing schedule is used.
For semantic segmentation, we use the pretrained KAConvNets from the image classification experiments as the backbones of PSPNet \cite{zhao2017pyramid}. The initial learning rate is set to 1e-3, with warm-up and cosine annealing applied for adjustment. Models are trained for 50 epochs using the AdamW optimizer.
In both object detection and segmentation experiments, pretrained models are fine-tuned with all backbone layers unfrozen. The batch size is set to 64.

All experiments are implemented in PyTorch and trained on eight NVIDIA A100 GPUs (80 GB each). 

\subsection{Image Classification}
We evaluate the performance of our KAConvNet architecture on the ImageNet-1K classification dataset across three parameter scales: KAConvNet-S (small), KAConvNet-B (medium), and KAConvNet-L (large).
Comprehensive comparisons are conducted against three categories of baseline models: traditional CNNs (e.g., RedNet \cite{li2021involution}, MobileOne \cite{vasu2023mobileone}, GhostNet \cite{han2020ghostnet}, ShuffleNet \cite{ma2018shufflenet}, and RevBiFPN \cite{chiley2023revbifpn}), Transformers (e.g., EfficientFormer \cite{li2022efficientformer}, GC-ViT \cite{hatamizadeh2023global}, DeiT-Tiny+ \cite{touvron2021going}, and LocalViT-T2T \cite{li2021localvit}), Mamba-based models (e.g., VMamba-T \cite{liu2024vmamba}, MambaVision \cite{hatamizadeh2025mambavision}) and KACNNs (e.g., KACN-VGG19 \cite{drokin2024kolmogorov}, KACN-VGG11 \cite{drokin2024kolmogorov}, JacobiKAN \cite{JacobiKAN2024}, Fast-KAN \cite{FastKAN2024}).

It is worth noting that due to the scale incompatibility between Mamba-based models and our proposed method, we adjusted the relevant settings to match the current model scale.  

As shown in \cref{table:imagenet1k}, our large-scale model KAConvNet-L achieves remarkable performance compared to other KA-based methods. Additionally, it attains competitive performance to mainstream Mamba-based models, CNNs, and ViTs.

For the medium-scale model KAConvNet-B, our model achieves a Top-1 accuracy of 76.8\% with 8.6M parameters and 1.4G FLOPs. This surpasses comparable models like GhostNetx1.3, RedNet-26, and DeiT-Tiny+ in both efficiency and accuracy.  

For the small-scale KAConvNet-S, we achieve a Top-1 accuracy of 73.7\% with 5.0M parameters and 0.7G FLOPs. Despite having slightly more parameters than lightweight CNNs like ShuffleNetV2 and RevBiFPN-S0, KAConvNet-S achieves significant accuracy gains by 1.5\%. It also outperforms transformer-based models like DeiT-Tiny and LocalViT-T2T with fewer resources.

These results highlight KAConvNet's ability to integrate convolution and KA mechanisms effectively, outperforming both mainstream Mamba-based models, CNNs and ViTs in accuracy and efficiency.

\subsection{Object Detection}
To evaluate the versatility of KAConvNet, we test its performance on the object detection task using the RTMDet-tiny detector \cite{lyu2022rtmdet} as the base framework. By replacing the backbone with KAConvNet variants (small, medium, and large) and comparing them with other mainstream models of similar scale, we measure mean Average Precision (mAP) on the MS COCO validation set. All models are trained using the mmdetection library \cite{mmdetection} with consistent settings and an input resolution of 640×640. As shown in \cref{tab:detection}, KAConvNet achieves mAP scores of 43.3, 45.8, and 48.0 for the small, medium, and large scales, respectively. 
Our KAConvNets are highly competitive compared to the excellent CNN-based and Transformer-based methods under the same object detection framework. 
These results demonstrate the effectiveness of KAConvNet in object detection.
\begin{table}[!h]
\centering
\footnotesize
\caption{Object detection results on MS COCO.}
\setlength{\tabcolsep}{7.25mm}
\begin{tabular}{lccc}
\hline
Models          & mAP  & mAP$_{50}$      & mAP$_{75}$ \\ \hline
EfficientNet-B2 \cite{tan2019efficientnet} &  47.6    &  65.3    &        51.4              \\
MobileOne-S4 \cite{vasu2023mobileone}    &   47.5   & 65.1         &      51.4             \\
IPT-T \cite{fu2022incepformer}           & 47.9     &  65.5         &    51.8              \\
HVT Base \cite{fein2024hvt}              &  47.8    &  65.6        &    51.5       \\
\textbf{KAConvNet-L (Ours)}     & \textbf{48.0}     &  \textbf{65.7}         & \textbf{51.9}        \\    \hline
MobileVit-S \cite{mehta2021mobilevit}    & 45.7     &  63.0         &   49.5               \\
PVT-Tiny \cite{wang2021pyramid}       & 45.7 &    62.8        &   49.5             \\
\textbf{KAConvNet-B (Ours)}     & \textbf{45.8} &   \textbf{63.1}         &   \textbf{49.6}                \\ \hline
MobileOne-S0 \cite{vasu2023mobileone}   &  43.1    &    60.4         &       47.0         \\
DeiT-Tiny \cite{touvron2021going}      & 43.0     &     60.5         &        46.8       \\
\textbf{KAConvNet-S (Ours)}    & \textbf{43.3}    & \textbf{60.6}           & \textbf{47.1}                \\ \hline
\end{tabular}
\label{tab:detection}
\end{table}

\begin{table}[!h]
\centering
\footnotesize
\caption{Semantic segmentation performance on Cityscapes.}
\setlength{\tabcolsep}{8mm}
\begin{tabular}{lcc}
\hline
Backbone               & Mean IoU & Mean Pixel Acc \\ \hline
ResNet-18 (baseline)   & 68.02    & 77.90          \\ \hline
RepVGG-A2 \cite{ding2021repvgg}  & 69.94    & 78.89 \\
MobileOne-S4 \cite{vasu2023mobileone} & 70.21   & 79.36  \\
HVT Base \cite{fein2024hvt}  & 70.32   & 79.25  \\
\textbf{KAConvNet-L (Ours)} & \textbf{70.58}    & \textbf{79.51}          \\ \hline
RepVGG-A1 \cite{ding2021repvgg}  & 68.52    & 78.33 \\
GhostNet1.3 \cite{han2020ghostnet} & 68.66   & 78.47 \\ 
LocalViT-T \cite{li2021localvit}  & 69.07   & 78.55  \\
ViL-Tiny-APE \cite{zhang2021multi}  & 69.10    & 78.52   \\
\textbf{KAConvNet-B (Ours)} & \textbf{69.20}    & \textbf{78.70}          \\ \hline
MobileV3-L 0.75 \cite{howard2019searching}       & 65.19    & 76.01          \\
DeiT-Tiny \cite{touvron2021going}  & 64.95   &       75.90\\ 
\textbf{KAConvNet-S (Ours)} & \textbf{65.32}    & \textbf{76.06}          \\ \hline
\end{tabular}
\label{tab:seg}
\vspace{-1mm}
\end{table}

\subsection{Semantic Segmentation}

Next, we assess KAConvNet on the semantic segmentation task using the Cityscapes dataset. We integrate KAConvNet into PSPNet \cite{zhao2017pyramid} and compare it with mainstream backbones across three parameter scales, measuring Mean Intersection over Union (Mean IoU) and Mean Pixel Accuracy (Mean Pixel Acc). All models are trained under the same settings, and the results are summarized in \cref{tab:seg}. 
Our KAConvNets outperform other competitive CNN-based and Transformer-basedIntern models.
Impressively, KAConvNet-L achieves the best performance, with a Mean IoU of 70.58 and a Mean Pixel Accuracy of 79.51, marking a significant improvement over established baselines such as ResNet-18.

These results confirm KAConvNet's robustness across classification, detection, and segmentation tasks, highlighting its effectiveness not only in image-level recognition but also in pixel-level tasks. This underscores its potential as a versatile backbone for a wide range of vision applications.

\begin{table}[!h]
\centering
\footnotesize
\caption{Accuracy results of different learnable activation functions $\tilde{\phi}$ and different operations $op$ on the CIFAR-100 dataset.}
\vspace{-1mm}
\setlength{\tabcolsep}{4mm}
\begin{tabular}{lccccc}
\hline
Method          & $\tilde{\phi} $     & $n$     & Acc (\%)     & \#Param.      & FLOPs           \\ \hline
VGG11 (baseline) & ---    & ---  & 69.74  & 29.0M  & 173.0M \\
KAVGG11        & PReLU    & 2 & 71.29  & 39.7M   & 369.3M  \\
KAVGG11        & B-Spline & 3 & 70.46  & 40.1M   & 366.5M  \\
KAVGG11        & B-Spline & 5 & 69.51  & 40.3M   & 366.5M  \\
KAVGG11        & GLinear  & 2 & 71.40  & 39.8M   & 371.2M  \\
KAVGG11        & GLinear  & 4 & 72.38  & 39.8M   & 371.2M  \\
KAVGG11        & GLinear  & 6 & 72.28  & 39.9M   & 371.2M  \\
KAVGG11        & GLinear  & 8 & 71.65  & 40.0M   & 371.2M  \\ \hline
\end{tabular}
\label{tab:abdifferentact}
\vspace{-1mm}
\end{table}

\subsection{Ablation Studies}

To further evaluate the effectiveness of our KAConvLayer, we conduct ablation experiments using VGG-like networks on the CIFAR-100 dataset. VGG-like networks, consisting solely of stacked convolution layers without residual structure, offer a straightforward comparison between the performance of our KAConvLayers and standard convolution layers. Specifically, we create KAVGG11 by replacing all standard convolution layers in VGG11 with KAConvLayers.

We first explore the impact of different learnable activation functions and their interval divisions on model performance. We test GLinear, PReLU, and B-spline functions, combining them with the basic activation function in KAVGG11 on CIFAR-100. The B-spline function is set to an order of 3, matching the configuration in KAN.

As shown in \cref{tab:abdifferentact}, most learnable activation functions enhance the model's performance, but the B-spline function results in lower accuracy. This overly complex activation function appears prone to overfitting, making it unsuitable for computer vision tasks.

For the GLinear function, we observe that increasing the number of intervals from 2 to 4 improves model accuracy by about 1\%. Performance stabilizes when the number of intervals is increased from 4 to 6, but accuracy declines by approximately 0.6\% when the number of intervals reaches 8. These findings indicate that finer interval divisions do not necessarily lead to better performance. The optimal number of intervals for a learnable activation function appears to depend on the complexity of the visual task.

\begin{table}[!h]
\centering
\footnotesize
\caption{Results of replacing the standard convolution layers in VGG11 with the KAConvLayers on the CIFAR-100.}
\vspace{-1mm}
\setlength{\tabcolsep}{3.25mm}
\begin{tabular}{lcccc}
\hline
Layers with KAConvLayer & Top-1 Acc & Top-5 Acc  & \#Param.      & FLOPs \\ \hline
\textbf{\ \textendash}             & 69.74      & 89.96   & 29.0M  &  173.0M  \\
$[1,2]$                 & 71.95      & 90.25    & 29.2M   & 204.5M   \\
$[3,4]$                 & 72.85      & 91.26    & 29.9M   & 246.9M   \\
$[5,6]$                 & 70.38      & 89.61    & 32.9M   & 241.8M \\
$[7,8]$                 & 67.65      & 88.67    & 34.7M   & 195.7M   \\
$[1,2,3,4]$             & 73.41      & 91.47    & 38.6M   & 265.5M  \\
$[5,6,7,8]$             & 68.96      & 89.03    & 29.8M   & 277.8M \\
$[1,2,3,4,5,6,7,8]$     & 71.40      & 89.86    & 39.8M   & 371.2M  \\ \hline
\end{tabular}
\label{tab:abkaconvlayer}
\vspace{-1mm}
\end{table}

Next, we investigate the impact of the number and position of KAConvLayers in the VGG11 architecture. For example, replacing the first and second standard convolution layers with KAConvLayers is denoted as "$[1,2]$". 
The notation "$-$" refers to the original VGG11 without KAConvLayers, and the deepest possible replacement occurs at the eighth layer, denoted as "$[1,2,3,4,5,6,7,8]$".

As demonstrated in \cref{tab:abkaconvlayer}, placing KAConvLayers in the shallow layers of the model yields better performance when the residual structure is not used. In contrast, replacing the deepest layers with KAConvLayers leads to reduced accuracy. Notably, the model without KAConvLayers in the last two layers shows improved performance. The best results are achieved when KAConvLayers are placed in the first four layers, resulting in an accuracy of 73.41\%.

\begin{table}[!h]
\centering
\footnotesize
\setlength{\tabcolsep}{7mm}
\caption{Ablation experiment with the residual structure. The KAConvLayers in the deep layers don't lead to poor performance.}
\label{tab: ab_for_res}
\begin{tabular}{lccc}
\hline
Stages \textit{w/} KAConvLayer & \#Params. & FLOPs & Acc (\%)  \\ \hline
---                     & 11.0M     & 1.7G  & 77.1 \\
$[1, 2]$                & 13.2M     & 2.1G  & 79.1 \\
$[3, 4]$                 & 15.3M     & 2.5G  & 79.4 \\
$[1, 2, 3, 4]$           & 17.5M     & 2.9G  & 80.1  \\ \hline
\end{tabular}
\end{table}

We further investigated the impact of KAConvLayers at different network depths within a residual structure. Using the KAConvNet architecture, we replaced the KAConvLayers at various stages with standard convolution layers and evaluated performance on the ImageNet-1K dataset.

As shown in \cref{tab: ab_for_res}, incorporating KAConvLayers alongside residual connections does not result in a significant performance drop in the deeper stages of the network. This suggests that the residual structure effectively mitigates the degradation observed when KAConvLayers are applied in deeper layers, helping to reduce harmful representations and maintain stable performance.

These experiments provide valuable insights into the optimal configuration of KAConvLayers, suggesting that both the choice of activation function and the placement of KAConvLayers play crucial roles in improving model performance.

\subsection{Computational Complexity Analysis}

We provide pseudo code of our KAConvLayer as follows:
\begin{lstlisting}[basicstyle=\footnotesize,frame=single]
def convkan:
    kaconv1_feature = conv1(act1(x))
    unfold_feature = unfold_and_rearrange(x)
    unfold_feature = rearrange(unfold_feature)
    kaconv2_feature = conv2(act2(unfold_feature))
    kaconv2_feature = rearrange(kaconv2_feature)
    kaconv2_feature = kaconv2_feature.reshape
    return kaconv1_feature * kaconv2_feature

def KAConvolutionLayer:
    kaconv_feature = convkan(x)
    kaconv_feature = norm(kaconv_feature)
    kaconv_feature = act(kaconv_feature)
    kaconv_feature = group_mlp(kaconv_feature)
        
    return mlp(kaconv_feature)
\end{lstlisting}

The computational complexity of a regular convolution layer is $\mathcal{O}(H_{out} \times W_{out} \times C_{out} \times C_{in} \times K \times K)$, while the computational complexity of our proposed KAConvLayer is $\mathcal{O}(H_{out} \times W_{out} \times C_{in} \times (2K^2+1) \times K^2 + H_{out} \times W_{out} \times C_{in} \times C_{out})$, where $H_{out}$ and $W_{out}$ denote the output height and width, $C_{in}$ and $C_{out}$ represent the input and output channel dimensions and $K$ is the kernel size.
When the number of output channels is small, the regular convolution computation is less than our KAConvLayer. As the number of output channels increases to a certain extent, our KAConvLayer requires less computation than regular convolution.

\section{Discussions and Limitations}
\subsection{Discussions}

\begin{table}[!h]
\centering
\footnotesize
\setlength{\tabcolsep}{15mm}
\caption{Comparison on GPU latency with KA-based convolution layers.}
\begin{tabular}{l|c}
\hline
                            & GPU Latency (ms)    \\  \hline
KAConvLayer        & 86.4  $\pm$ 2.9             \\
CKAN-Layer  \cite{bodner2024convolutional}                      & 249.8  $\pm$ 4.6             \\
KACN-Layer  \cite{drokin2024kolmogorov}                       & 101.5  $\pm$ 3.1             \\
Fast-KAN-Layer  \cite{FastKAN2024}                   & 99.7  $\pm$ 1.9 
\\
JacobiKAN-Layer  \cite{JacobiKAN2024}                  & 114.5  $\pm$ 4.3 
\\
\hline
\end{tabular}
\label{table: GPU_Latency}
\end{table}

We compared the inference latency of individual layers among different KA-based convolution layers, with the tensor dimension set to $32\times 256\times 64\times 64$ and kernel size fixed at $3\times 3$. The inference latency results tested on the H800 GPU are presented in \cref{table: GPU_Latency}.
Our method achieves superior inference speed compared to other competing methods, which can be attributed to the linearly learnable GLinear activation function and the channel-wise processing mechanism adopted in our design. This experimental result further verifies the superiority of the proposed method in terms of inference efficiency. 

\begin{figure}[!htb]
\centering
\subfloat[\fontsize{8}{5}\selectfont Image]{\includegraphics[width = 0.225\textwidth]{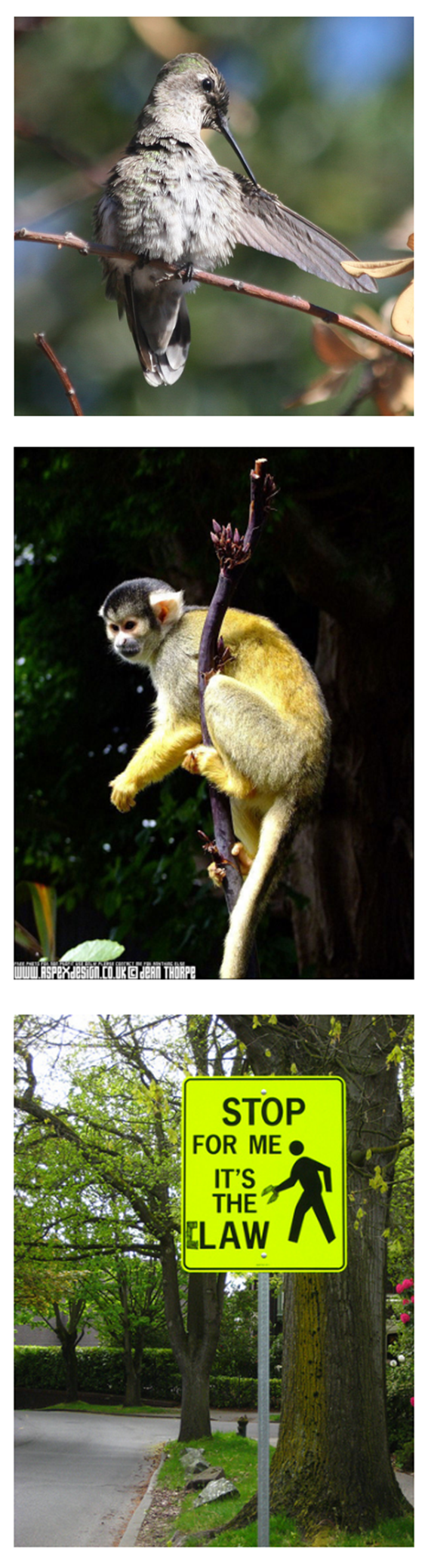}\label{subfig:oriimg}}
\subfloat[\fontsize{8}{5}\selectfont ConvNeXt-T]{\includegraphics[width = 0.225\textwidth]{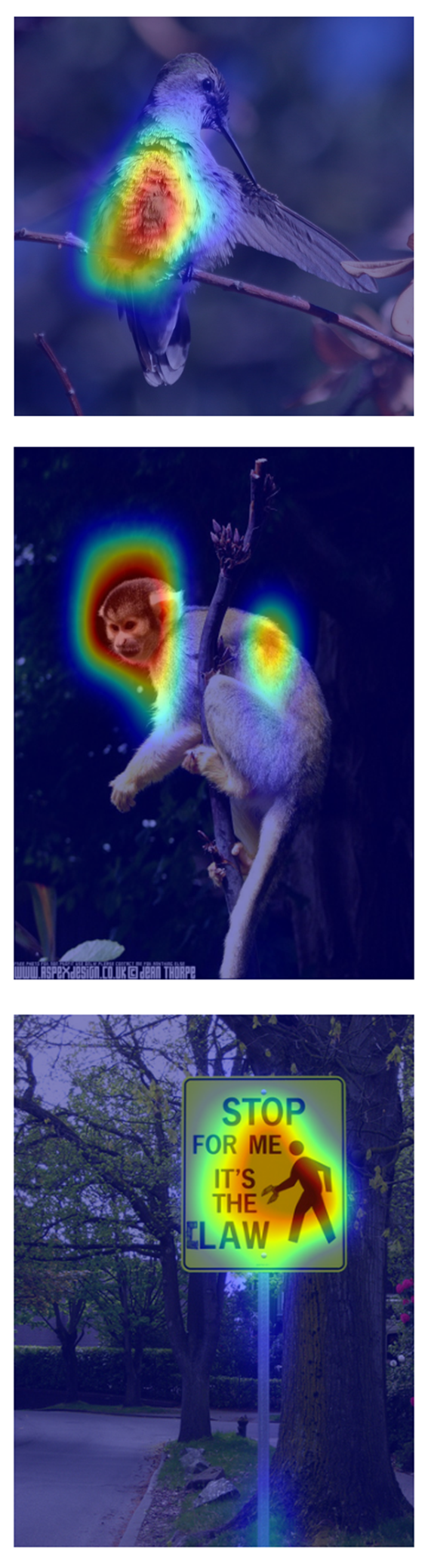}\label{subfig:convnextcam}}
\subfloat[\fontsize{8}{5}\selectfont ConvNet-L$^{\dagger}$]{\includegraphics[width = 0.225\textwidth]{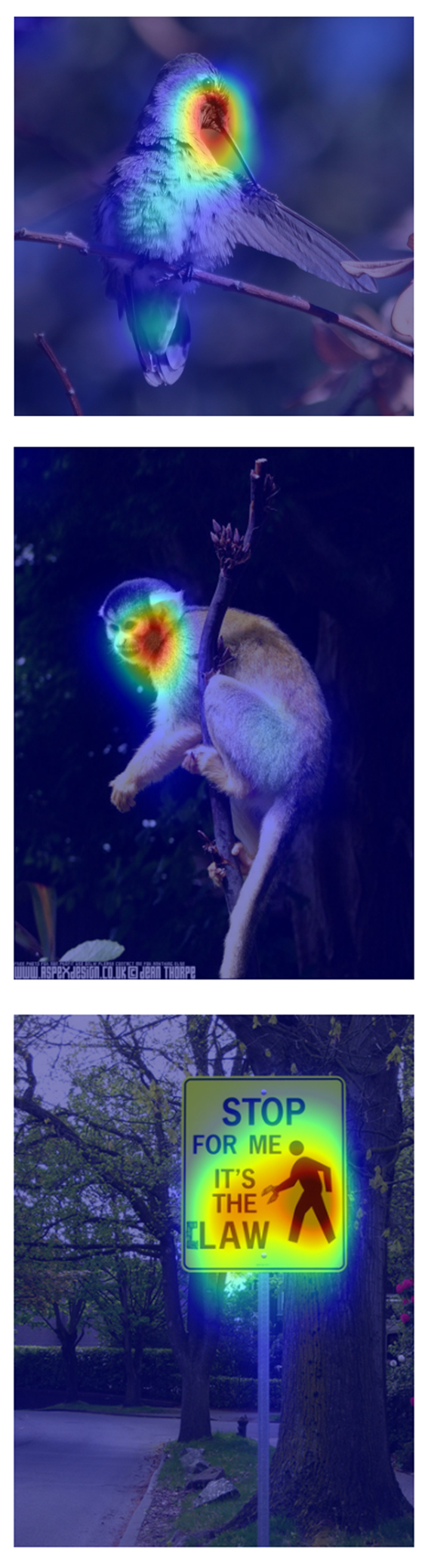}\label{subfig:convnetcam}}
\subfloat[\fontsize{8}{5}\selectfont KAConvNet-L]{\includegraphics[width = 0.225\textwidth]{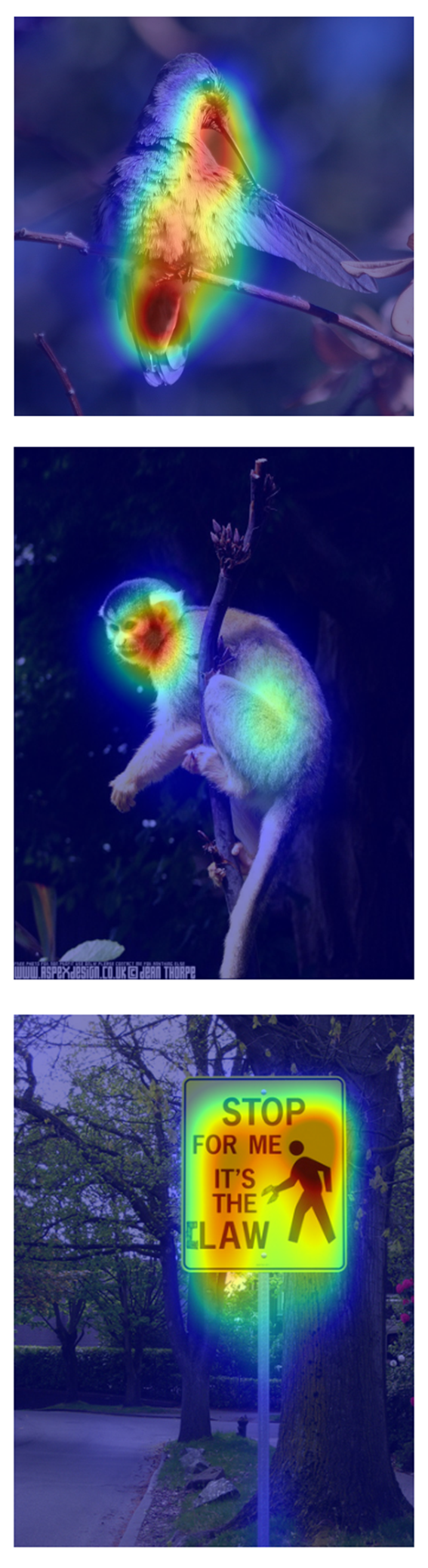}\label{subfig:kaconvnetcam}}
\caption{Grad-CAM of ConvNeXt-T, ConvNet-L$^{\dagger}$ and our KAConvNet-L. }
\label{fig:cam}
\vspace{-4mm}
\end{figure}
\begin{table}[!h]
\footnotesize
\centering
\caption{Comparison results of KAConv Layer and standard convolution layer in the same architecture. $^{\dagger}$ indicates replacing all KAConvLayers in the KAConvNet architecture with standard convolution layers configured identically. $^{\prime}$ signifies expanding the $3\times 3$ KAConv layer in the first stage of KAConvNet to a $5\times 5$ configuration.}
\vspace{2mm}
\setlength{\tabcolsep}{3mm}
\begin{tabular}{lcccc}
\hline
Method      & \#Params. & FLOPs  & Inference Speed & Acc (\%)  \\ \hline
ConvNet-L$^{\dagger}$   & 11.0M  & 1.7G      &  15.1ms          & 77.1     \\
KAConvNet-L             & 17.5M  & 2.9G      &  20.8ms          & 80.1     \\
KAConvNet-L$^{\prime}$  & 17.8M  & 4.0G       & 21.0ms           & 80.3     \\ \hline
ConvNet-B$^{\dagger}$   & 5.6M   & 0.7G      &  14.9ms          & 74.0     \\
KAConvNet-B             & 8.6M   & 1.4G       &  20.6ms          & 76.8     \\
KAConvNet-B$^{\prime}$  & 8.9M   & 2.0G       & 20.8ms      & 77.0     \\ \hline
ConvNet-S$^{\dagger}$   & 3.5M   & 0.4G     &  8.5ms         & 70.1     \\
KAConvNet-S             & 5.0M   & 0.7G       &  11.4ms          & 73.7     \\
KAConvNet-S$^{\prime}$  & 5.1M   & 1.1G       & 11.5ms           & 74.2     \\ \hline
\end{tabular}
\label{tab:discuss}
\vspace{-2mm}
\end{table}

We present Gradient-weighted Class Activation Mapping (Grad-CAM) visualizations for ConvNeXt-T \cite{liu2022convnet} (82.1\% accuracy on ImageNet-1K), ConvNet-L$^{\dagger}$\footnote{$^{\dagger}$ indicates that all KAConvLayers in the KAConvNet architecture are replaced with standard convolution layers, configured identically.}, and KAConvNet-L, as shown in \cref{fig:cam}.

The results reveal that, compared to ConvNeXt-T and ConvNet-L, KAConvNet-L more effectively and comprehensively locates objects in the images. For instance, in the hummingbird image, KAConvNet-L accurately highlights the entire bird, including the beak and tail. Similarly, in the street sign image, KAConvNet-L not only identifies the entire signal board, but also focuses more precisely on the text and logo, demonstrating a more detailed and accurate object attention.

Our experiments show that while KAConvNet performs well on the ImageNet-1K dataset, it still slightly lags behind existing large-kernel convolution networks. We attribute this performance gap to the small kernel size ($3 \times 3$) in the KAConvLayer, which limits the effective receptive field and hinders the model's ability to capture long-range pixel dependencies.

To validate this hypothesis, we increase the kernel size to $5 \times 5$ in the first stage of a pretrained KAConvNet and train the model for 50 epochs. Additionally, we examine the performance differences between KAConvLayer and standard convolution within the same modern convolutional neural network architecture.

As shown in \cref{tab:discuss}, replacing standard convolutional layers with KAConvLayers leads to a significant improvement in accuracy. Although the use of a $5\times 5$ KAConvLayer increases computational cost compared to the $3\times 3$ version, it results in only a slight rise in parameter count and inference time. these results demonstrates the scalability of our KAConvnet to some extent.

Previous studies \cite{ding2022scaling,liu2022more,ding2024unireplknet} have highlighted that shallow networks with large convolution kernels often outperform deeper models with smaller kernels. We believe that combining KAConvLayer with larger convolution kernels is a promising direction for future research.

\cref{tab:abdifferentact} also shows that the most of learnable activation functions used on the weights enable the model to achieve accuracy gains. The optimal number of intervals for a learnable activation function appears to depend on the complexity of the visual task.

Finally, we advocate that using KAConvLayer together with the residual structure like KAConvNet, or only using KAConvLayer in the shallow layer of the network, in the hope that the model will achieve better performance after training.

\subsection{Limitations}
Since learnable activation functions for feature map need to be placed on weights, it inevitably leads to an increase in the amount of computation. We use the unfold function, linear learnable activation function and channel-separable rectangular convolution to maximize the acceleration of this process, but the computational efficiency is still somewhat lower than that of the standard convolution.

We acknowledge that we currently lack an effective solution to accelerate the runtime overhead induced by segmented activation functions—a limitation that is, in fact, a common drawback of existing KAN-based architectures. It is important to clarify that the core contribution of this work is to propose a KAN-based convolution method that achieves significant improvements over KAN methods in both accuracy and inference speed.
For future work, advancing KAN-based convolution methods to match the efficiency of leading ViTs and CNNs will require more explorative experiments on architectural designs, as well as in-depth research and development of mathematical theorems underpinned by the KA representation theorem. 

\section{Conclusion}
In this paper, we apply the Kolmogorov–Arnold representation theorem to 2D convolution, introducing the KAConvLayer, which demonstrates both strong fitting ability and method interpretability. Building on this, we design a network architecture called KAConvNet, which has shown excellent performance in classic computer vision tasks.
On the one hand, our results highlight the superior fitting ability of the KAConvLayer compared to standard convolution layers. 
On the other hand, we propose using residual structures to increase the depth of KAConvNets, thereby preventing performance degradation.
Looking ahead, our future work will focus on leveraging KAConvLayers with larger kernels and exploring computational acceleration techniques for KA convolutions. 
We believe our work offers a fresh perspective on convolutional methods for the 2020s.




\bibliographystyle{elsarticle-num.bst}
\bibliography{mybibfile}


\clearpage
\appendix
\section{Visualization Results in Semantic Segmentation}
\begin{figure*}[!ht]
    \centering
    \includegraphics[width=0.95\linewidth]{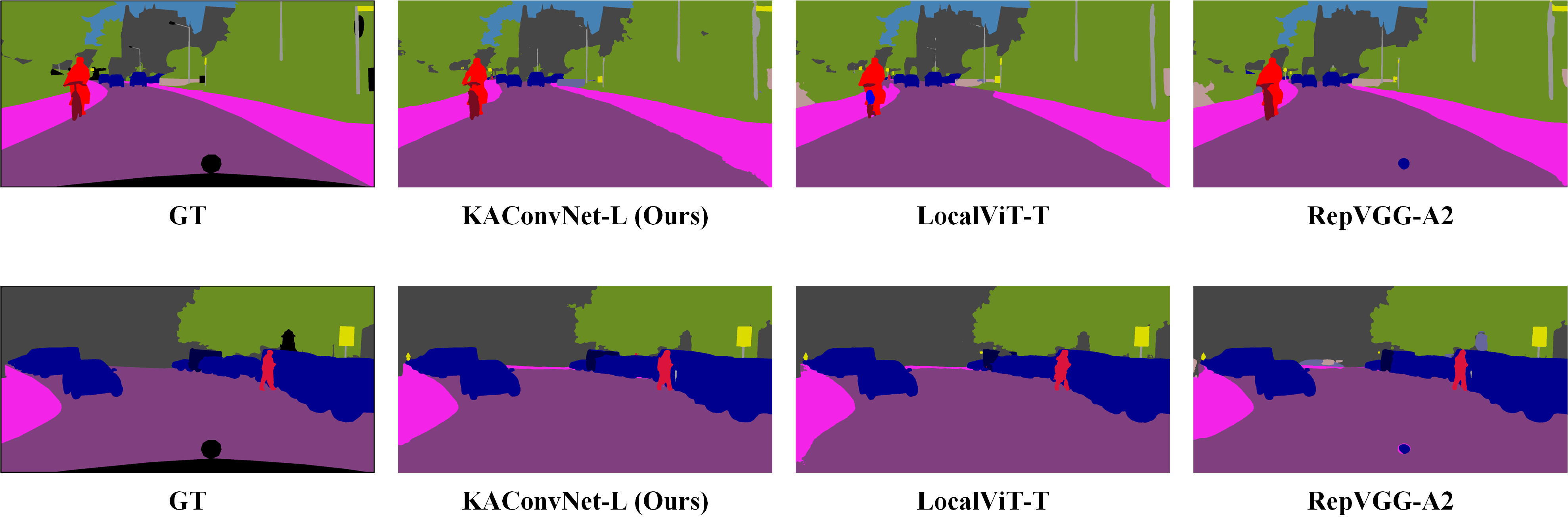}
    \caption{Visual comparison of semantic segmentation performance on Cityscapes.}
    \label{fig:visual_semseg}
\end{figure*}

As shown in \cref{fig:visual_semseg}, our KAConvNet demonstrates superior capability in accurately delineating object boundaries.

\end{document}